\documentclass[11pt,a4paper]{article}

\usepackage[T1]{fontenc}
\usepackage[utf8]{inputenc}
\usepackage{lmodern}
\usepackage[margin=2.5cm]{geometry}
\usepackage{amsmath,amssymb}
\usepackage{graphicx}
\usepackage{booktabs}
\usepackage{array}
\usepackage{multirow}
\usepackage{float}
\usepackage{caption}
\usepackage{subcaption}
\usepackage{hyperref}
\usepackage{xcolor}
\usepackage{microtype}
\usepackage{parskip}
\usepackage{enumitem}
\usepackage{url}
\usepackage{natbib}
\usepackage{fancyhdr}
\usepackage{titlesec}
\usepackage{abstract}

\hypersetup{
  colorlinks=true,
  linkcolor=blue!70!black,
  citecolor=blue!70!black,
  urlcolor=blue!70!black,
  pdftitle={Benchmarking Large Language Models on Multi-Sensor Physical Hazard Assessment},
  pdfauthor={Faizan Iqbal}
}

\titleformat{\section}{\large\bfseries}{\thesection.}{0.5em}{}
\titleformat{\subsection}{\normalsize\bfseries}{\thesubsection}{0.5em}{}

\newcommand{\Em}{E_m}
\newcommand{\ci}{C_i}
\newcommand{\li}{L_i}

\begin{document}

% ── Title block ────────────────────────────────────────────────────────────────
\begin{titlepage}
\centering
\vspace*{2cm}

{\LARGE\bfseries Benchmarking Large Language Models on\\[6pt]
Multi-Sensor Physical Hazard Assessment\par}

\vspace{1.5cm}

{\large
\textbf{Faizan Iqbal}\\[4pt]
Lovely Professional University, India\\[4pt]
\href{https://orcid.org/0009-0002-8998-9347}{\texttt{orcid.org/0009-0002-8998-9347}}\\[6pt]
\href{https://huggingface.co/faizaniqbal}{\texttt{huggingface.co/faizaniqbal}}
\\
\href{https://x.com/faizaniqbal__52}{\texttt{x.com/faizaniqbal\_\_52}}
\par}

\vspace{1cm}
{\small May 2026}

\vspace{2cm}

\begin{abstract}
\noindent
We present an empirical benchmark evaluating how five large language models
assess multi-sensor physical hazard data. Testing 60 scenarios across three
categories---multi-sensor joint assessment, response proportionality, and
pattern disambiguation---with 1{,}800 API calls at temperature~0.0, we find
that all tested models consistently produced no precautionary warning signal
across the tested scenarios where multiple sensors are simultaneously elevated
below their individual safety limits, while achieving near-perfect accuracy on
single-sensor threshold violations. All five models (ChatGPT-4o, Gemini~2.5
Flash, DeepSeek, Kimi, Llama~3.1~8B) score near zero on Category~A
multi-sensor scenarios ($Q_2$: 0.000--0.208; $Q_3$: 0.000--0.592) compared
to strong performance on single-sensor scenarios (Category~B $Q_1$:
0.975--1.000). Structured tabular formatting shows no consistent advantage
over plain prose; ChatGPT-4o performs significantly better under prose
($p=0.001$). These findings have direct implications for practitioners
deploying the tested models in physical safety monitoring systems.
\end{abstract}

\end{titlepage}

% ══════════════════════════════════════════════════════════════════════════════
\section{Introduction}

Large language models are increasingly considered for deployment in
safety-critical physical monitoring contexts including industrial environmental
sensors, building automation systems, and occupational health applications. In
such systems, LLMs receive streams of numerical sensor readings and are
expected to identify threshold violations, classify hazard types, and recommend
proportionate actions.

A critical and underexplored capability requirement is the assessment of
conditions where multiple sensors are simultaneously elevated below their
individual safety limits. Occupational hygiene standards address this through
the additive exposure index~\citep{osha1910}:
\begin{equation}
  \Em = \frac{C_1}{L_1} + \frac{C_2}{L_2} + \cdots + \frac{C_n}{L_n}
  \label{eq:osha}
\end{equation}
where $\Em > 1.0$ indicates combined exposure concern even when no individual
$\ci$ exceeds $\li$. Real-world examples include poorly ventilated occupied
spaces where CO, CO$_2$, VOC, and PM$_{2.5}$ are each near but below their
individual limits. Whether current LLMs respond appropriately to such
conditions---by flagging concern and recommending precautionary action---has
not been systematically studied.

This paper addresses four research questions:
\begin{enumerate}[leftmargin=*,label=\textbf{RQ\arabic*.}]
  \item Do tested models produce appropriate precautionary signals when multiple
        sensors are simultaneously elevated below individual safety limits?
  \item Do recommended actions scale proportionally with threshold exceedance
        magnitude?
  \item Can tested models correctly identify hazard type from multi-sensor
        patterns?
  \item Does structured tabular input formatting improve assessment quality
        compared to equivalent prose?
\end{enumerate}

% ══════════════════════════════════════════════════════════════════════════════
\section{Related Work}

\subsection{LLM Reasoning Benchmarks}

Standard LLM benchmarks including MMLU~\citep{hendrycks2021},
BIG-Bench~\citep{srivastava2022}, and GSM8K~\citep{cobbe2021} evaluate
mathematical and commonsense reasoning but do not address physical sensor
interpretation grounded in real regulatory safety standards. Physical AI
benchmarks~\citep{xu2024penetrative} test spatial and physical common sense but
not numerical threshold reasoning against occupational health limits.

\subsection{LLM IoT and Sensor Processing}

IoT-LLM~\citep{an2024iotllm} demonstrated that LLMs struggle with dense
numerical IoT data, achieving 49.4\% improvement with structured preprocessing
across five sensing tasks. SensorBench~\citep{sensorbench2025} benchmarked
LLMs on automated signal-processing coding tasks, finding human experts
outperform LLMs by over 60\% on complex multi-step tasks. Both works address
classification and signal-processing tasks rather than threshold-based safety
assessment against regulatory standards.

\subsection{LLM Safety Evaluation}

R-Judge~\citep{yuan2024rjudge} and SafetyBench~\citep{zhang2024safetybench}
evaluate LLM safety awareness in agent interaction and content safety contexts.
These benchmarks address behavioural safety of LLMs rather than their ability
to correctly interpret physical sensor measurements against numerical
thresholds.

\subsection{LLM Numerical Reasoning}

Research has documented LLM failures in numerical comparison tasks including
the widely cited $9.11 > 9.9$ error~\citep{zhao2024order}.
\citet{zhou2024larger} showed that larger models can become less reliable on
certain numerical tasks. HalluLens~\citep{bang2025hallulens} provides a
comprehensive hallucination taxonomy. Our work extends this to safety-critical
sensor contexts, finding that while all tested models handle individual
threshold arithmetic reliably, they consistently produce no precautionary
warning signal in multi-sensor elevated scenarios.

\subsection{Gap}

To the best of our knowledge, no prior work systematically benchmarks LLM
performance on multi-sensor physical hazard assessment using internationally
recognised safety standards as objective ground truth. This paper addresses
that gap.

% ══════════════════════════════════════════════════════════════════════════════
\section{Benchmark Design}

\subsection{Safety Thresholds}

All scenarios use the internationally recognised safety thresholds listed in
Table~\ref{tab:thresholds} as ground truth anchors.

\begin{table}[H]
\centering
\caption{Safety thresholds used as benchmark ground truth anchors.}
\label{tab:thresholds}
\begin{tabular}{llll}
\toprule
\textbf{Sensor} & \textbf{Threshold} & \textbf{Standard} \\
\midrule
Carbon monoxide (CO)           & 70 ppm      & NIOSH IDLH    \\
Carbon dioxide (CO$_2$)        & 2000 ppm    & ASHRAE 62.1   \\
Particulate matter PM$_{2.5}$  & 55~\textmu g/m$^3$ & WHO 2021 \\
Volatile organic compounds     & 1000 ppb    & ASHRAE 62.1   \\
Air temperature                & 43~\textdegree C  & OSHA heat stress \\
Relative humidity              & 90\%        & ASHRAE 55     \\
Sound pressure level           & 100 dB      & OSHA 1910.95  \\
Vibration acceleration         & 1.0 m/s$^2$ & ISO 2631-1    \\
\bottomrule
\end{tabular}
\end{table}

\subsection{Category A --- Multi-sensor Joint Assessment (20 scenarios)}

Twenty scenarios where no individual sensor exceeds its danger threshold, but
multiple sensors are simultaneously elevated. For scenarios involving chemical
sensors (CO, CO$_2$, VOC, PM$_{2.5}$), ground truth is established using the
OSHA additive exposure index (Equation~\ref{eq:osha}), where $\Em > 1.0$
indicates combined exposure concern for substances with similar physiological
effects. All chemical-sensor Category~A scenarios have $\Em > 1.0$ by
construction. For scenarios involving mixed sensor types, ground truth reflects
established occupational hygiene practice that simultaneous elevation of
multiple environmental stressors below individual limits warrants precautionary
action, consistent with ACGIH multi-stressor guidance~\citep{acgih2024}.

The expected model behaviour in Category~A is to flag precautionary concern
and recommend ventilation or investigation rather than declaring the environment
unambiguously safe. The primary outcome measures are $Q_2$ (hazard
classification) and $Q_3$ (action recommendation). $Q_1$ in Category~A
confirms individual sensor status as a sanity check.

Manual spot-checking of 30 randomly sampled Category~A responses (six per
model) confirmed that scorer output matches human judgement in all cases.

\subsection{Category B --- Proportionality Scenarios (20 scenarios)}

A single sensor exceeds its threshold at magnitudes from 1\%
(CO~=~70.7~ppm vs 70~ppm threshold) to 400\% (CO~=~350~ppm). Tests whether
recommended actions scale proportionally with exceedance severity. $Q_1$ is
the primary outcome measure.

\subsection{Category C --- Pattern Disambiguation Scenarios (20 scenarios)}

Tests correct hazard type identification from sensor patterns including
heat-not-fire, PM$_{2.5}$-not-fire, CO-leak-not-fire, ventilation failure,
structural hazard, and all-safe conditions.

\subsection{Scoring Rubric}

Three independent questions per scenario:

\textbf{$Q_1$ (Threshold Arithmetic):} Each sensor's measured value compared
to its threshold. Scored 1.0 for correct numeric comparisons, 0.0 for parrot
responses or numerically incorrect comparisons.

\textbf{$Q_2$ (Hazard Classification):} Is precautionary action needed? Scored
1.0 for correct verdict with sensor justification, 0.5 for correct verdict
only, 0.0 for incorrect verdict.

\textbf{$Q_3$ (Action Recommendation):} Scored against seven semantic action
classes (evacuate, ventilate, emergency services, cool environment, hydrate,
no action, report facilities). Score equals the fraction of required classes
present.

A question-echo artefact was identified and corrected during analysis; full
details are in Appendix~\ref{app:scorer}. All results use the corrected scorer.

% ══════════════════════════════════════════════════════════════════════════════
\section{Experimental Setup}

Five models were evaluated:
\begin{itemize}[noitemsep]
  \item ChatGPT-4o (OpenAI API, \texttt{gpt-4o})
  \item Gemini~2.5 Flash (Google API --- Gemini~2.0 Flash was retired by
        the provider before experiments ran)
  \item DeepSeek (\texttt{deepseek-chat})
  \item Kimi (\texttt{moonshot-v1-auto})
  \item Llama~3.1~8B Instant (Groq API)
\end{itemize}

All calls used temperature~$= 0.0$, \texttt{max\_tokens}~$= 600$. Two prompt
formats per scenario: \textbf{C6} (structured tabular with sensor, measured
value, threshold, and standard columns) and \textbf{C7} (plain prose,
identical information). Three independent runs per
scenario\,--\,model\,--\,format combination. Total: $60 \times 2 \times 5
\times 3 = 1{,}800$ API calls, zero errors, 234.4 minutes runtime.

An 8-scenario pilot study (B3--B10) was conducted prior to full benchmark
construction as a single-run exploratory analysis to validate the scoring
rubric and inform category design. It was not used to select scenarios based on
model performance, preventing data leakage into the benchmark design.

The benchmark scenarios, prompt templates, evaluation scorer, raw results, and
analysis scripts are released at
\url{https://github.com/Faizaniqbal52/PhysicalHazardBenchmark}~\citep{benchmark2026}.

% ══════════════════════════════════════════════════════════════════════════════
\section{Results}

\subsection{Q\textsubscript{1} Threshold Arithmetic Accuracy}

\begin{figure}[H]
\centering
\includegraphics[width=0.92\textwidth]{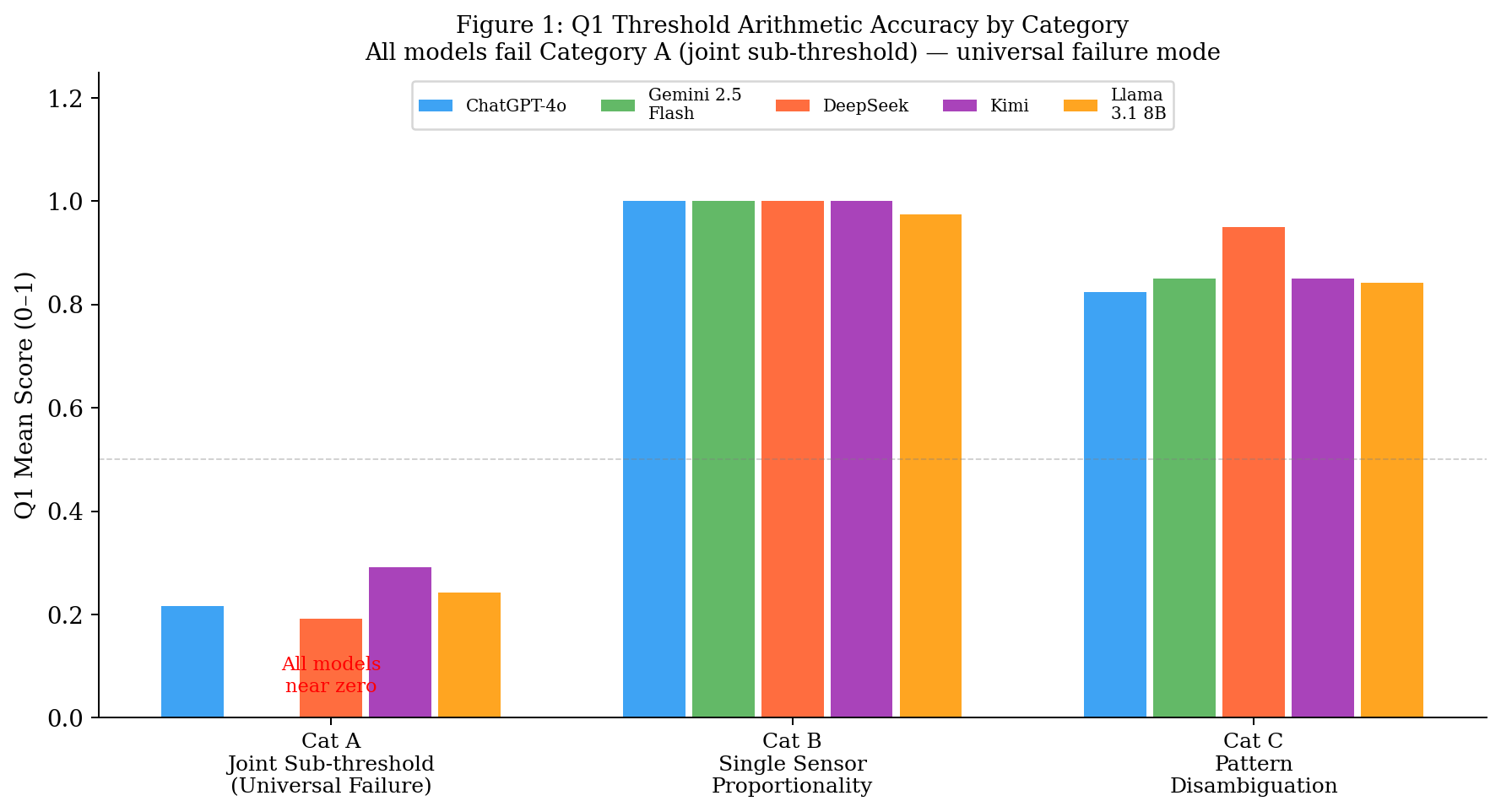}
\caption{$Q_1$ Threshold Arithmetic Accuracy by scenario category and model.
         Category~A (multi-sensor joint assessment) scores are substantially
         lower than Category~B (single-sensor proportionality) across all
         tested models.}
\label{fig:accuracy}
\end{figure}

\begin{figure}[H]
\centering
\includegraphics[width=0.95\textwidth]{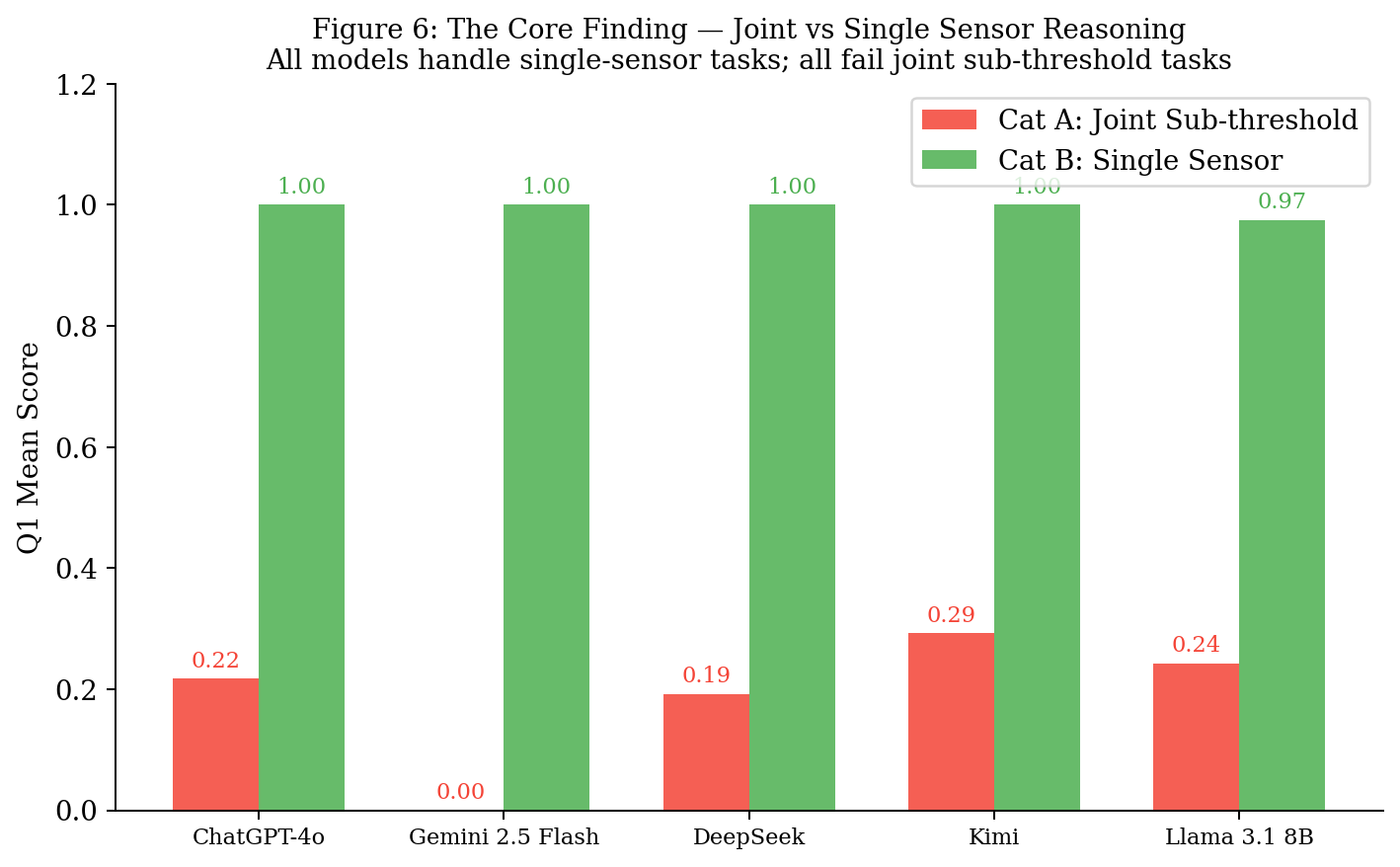}
\caption{Comparison of Category~A (multi-sensor joint assessment) and
         Category~B (single-sensor proportionality) $Q_1$ scores across
         tested models. Category~A scores are substantially lower than
         Category~B for all five models.}
\label{fig:core}
\end{figure}

Table~\ref{tab:q1} reports $Q_1$ accuracy by model and category.
Figure~\ref{fig:accuracy} visualises the category-level comparison, and
Figure~\ref{fig:core} summarises the contrast between Category~A and
Category~B performance.

\begin{table}[H]
\centering
\caption{$Q_1$ Threshold Arithmetic Accuracy by model and scenario category.
         For Category~A, $Q_1$ confirms individual sensor status (all below
         individual limits). Primary Category~A outcomes are $Q_2$ and $Q_3$
         (Tables~\ref{tab:q2} and~\ref{tab:q3}).}
\label{tab:q1}
\begin{tabular}{lcccc}
\toprule
\textbf{Model} & \textbf{Cat A} & \textbf{Cat B} & \textbf{Cat C} & \textbf{Overall} \\
\midrule
ChatGPT-4o       & 0.217 & 1.000 & 0.825 & 0.681 \\
Gemini 2.5 Flash & 0.000 & 1.000 & 0.850 & 0.617 \\
DeepSeek         & 0.192 & 1.000 & 0.950 & 0.714 \\
Kimi             & 0.292 & 1.000 & 0.850 & 0.714 \\
Llama 3.1 8B     & 0.242 & 0.975 & 0.842 & 0.686 \\
\bottomrule
\end{tabular}
\end{table}

\subsection{Category A Per-Scenario Detail}

\begin{figure}[H]
\centering
\includegraphics[width=0.95\textwidth]{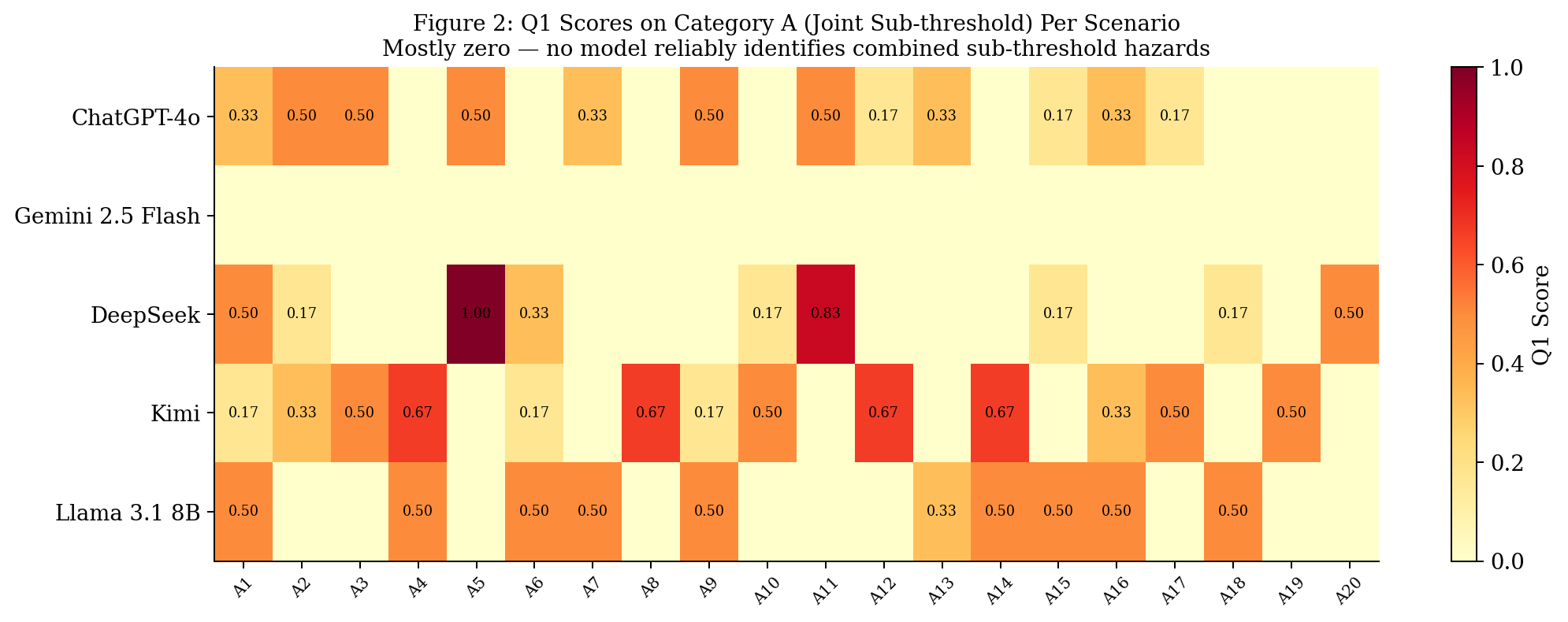}
\caption{Category~A $Q_1$ scores across all 20 scenarios and five tested
         models. Low scores across the matrix indicate that Category~A
         performance is consistently lower than Category~B, with variation
         across scenarios and models.}
\label{fig:heatmap}
\end{figure}

Table~\ref{tab:cata} shows $Q_1$ scores per Category~A scenario across all
five models. Figure~\ref{fig:heatmap} visualises the distribution.

\begin{table}[H]
\centering
\caption{$Q_1$ per Category~A scenario across tested models. Scores are
         consistently low, with no model achieving above 0.833 on any
         individual scenario and mean scores ranging from 0.000 to 0.292.}
\label{tab:cata}
\setlength{\tabcolsep}{5pt}
\begin{tabular}{lccccc}
\toprule
\textbf{Scenario} & \textbf{ChatGPT} & \textbf{Gemini} & \textbf{DeepSeek} & \textbf{Kimi} & \textbf{Llama} \\
\midrule
A1  & 0.333 & 0.000 & 0.500 & 0.167 & 0.500 \\
A2  & 0.500 & 0.000 & 0.167 & 0.333 & 0.000 \\
A3  & 0.500 & 0.000 & 0.000 & 0.500 & 0.000 \\
A4  & 0.000 & 0.000 & 0.000 & 0.667 & 0.500 \\
A5  & 0.500 & 0.000 & 1.000 & 0.000 & 0.000 \\
A6  & 0.000 & 0.000 & 0.333 & 0.167 & 0.500 \\
A7  & 0.333 & 0.000 & 0.000 & 0.000 & 0.500 \\
A8  & 0.000 & 0.000 & 0.000 & 0.667 & 0.000 \\
A9  & 0.500 & 0.000 & 0.000 & 0.167 & 0.500 \\
A10 & 0.000 & 0.000 & 0.167 & 0.500 & 0.000 \\
A11 & 0.500 & 0.000 & 0.833 & 0.000 & 0.000 \\
A12 & 0.167 & 0.000 & 0.000 & 0.667 & 0.000 \\
A13 & 0.333 & 0.000 & 0.000 & 0.000 & 0.333 \\
A14 & 0.000 & 0.000 & 0.000 & 0.667 & 0.500 \\
A15 & 0.167 & 0.000 & 0.167 & 0.000 & 0.500 \\
A16 & 0.333 & 0.000 & 0.000 & 0.333 & 0.500 \\
A17 & 0.167 & 0.000 & 0.000 & 0.500 & 0.000 \\
A18 & 0.000 & 0.000 & 0.167 & 0.000 & 0.500 \\
A19 & 0.000 & 0.000 & 0.000 & 0.500 & 0.000 \\
A20 & 0.000 & 0.000 & 0.500 & 0.000 & 0.000 \\
\bottomrule
\end{tabular}
\end{table}

\subsection{Format Effect}

\begin{figure}[H]
\centering
\includegraphics[width=0.85\textwidth]{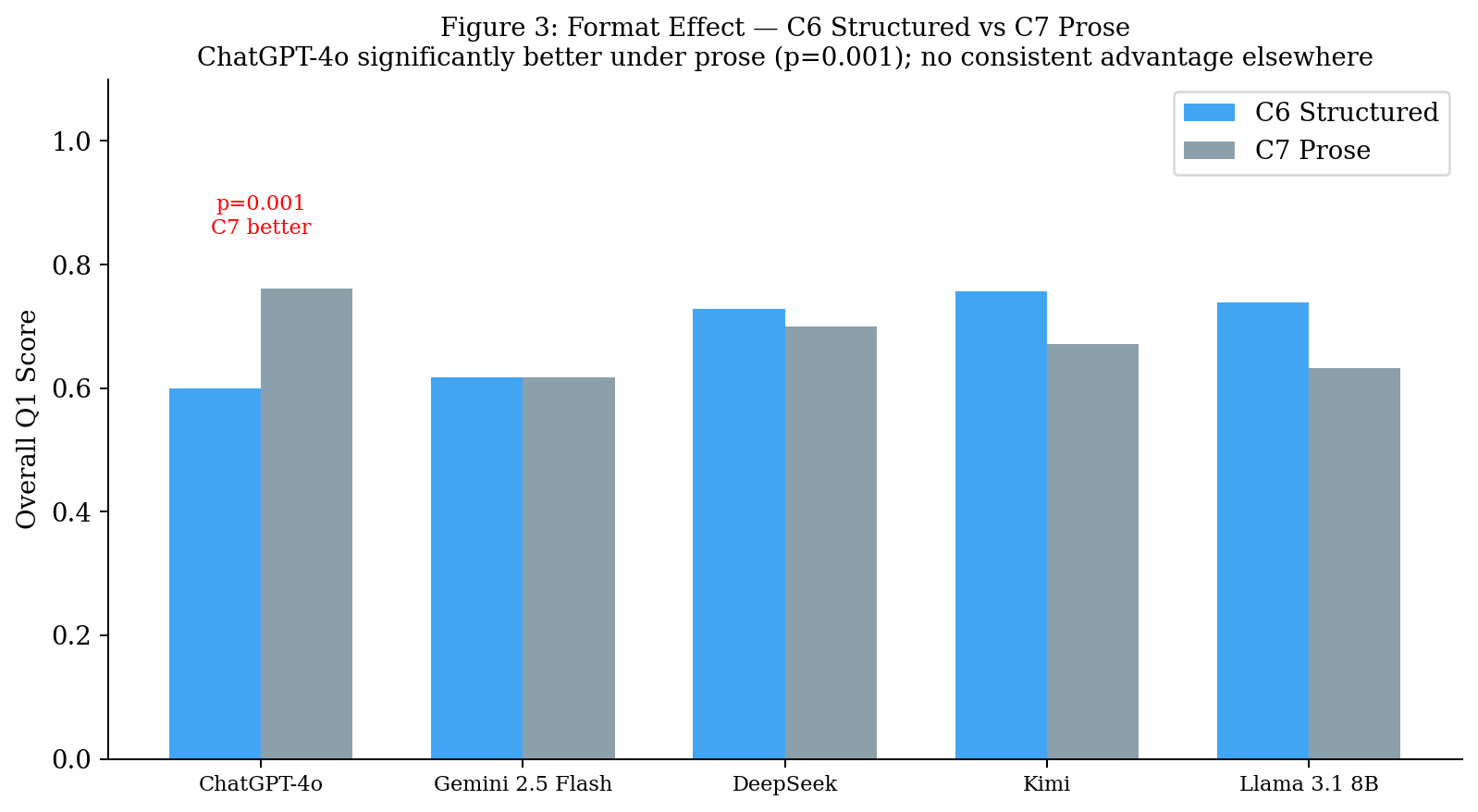}
\caption{Format effect: C6 (structured table) vs.\ C7 (plain prose) on
         overall $Q_1$. ChatGPT-4o scores significantly lower under structured
         format ($p=0.001$, $d=-0.64$). No other model shows a significant
         difference after Bonferroni correction.}
\label{fig:format}
\end{figure}

Table~\ref{tab:format} and Figure~\ref{fig:format} report the effect of
structured vs.\ prose formatting on overall $Q_1$.

\begin{table}[H]
\centering
\caption{Format effect on overall $Q_1$. Wilcoxon signed-rank test,
         two-sided, $N=60$ paired scenarios. Bonferroni-corrected
         significance threshold: $p < 0.010$.}
\label{tab:format}
\begin{tabular}{lcccc}
\toprule
\textbf{Model} & \textbf{C6 Struct.} & \textbf{C7 Prose} & \textbf{$\Delta$} & \textbf{$p$} \\
\midrule
ChatGPT-4o       & 0.600 & 0.761 & $-$0.161 & \textbf{0.001}$^*$ \\
Gemini 2.5 Flash & 0.617 & 0.617 & $ $0.000 & 1.000 \\
DeepSeek         & 0.728 & 0.700 & $+$0.028 & 0.559 \\
Kimi             & 0.756 & 0.672 & $+$0.084 & 0.064 \\
Llama 3.1 8B     & 0.739 & 0.633 & $+$0.106 & 0.106 \\
\bottomrule
\multicolumn{5}{l}{\small $^*$Significant after Bonferroni correction.}
\end{tabular}
\end{table}

\subsection{Q\textsubscript{2} Hazard Classification}

\begin{table}[H]
\centering
\caption{$Q_2$ Hazard Classification Accuracy. Scores are lower bounds due
         to rubric strictness (see Section~\ref{sec:discussion}).}
\label{tab:q2}
\begin{tabular}{lcccc}
\toprule
\textbf{Model} & \textbf{Cat A} & \textbf{Cat B} & \textbf{Cat C} & \textbf{Overall} \\
\midrule
ChatGPT-4o       & 0.008 & 0.242 & 0.342 & 0.197 \\
Gemini 2.5 Flash & 0.008 & 0.017 & 0.258 & 0.094 \\
DeepSeek         & 0.208 & 0.217 & 0.475 & 0.300 \\
Kimi             & 0.000 & 0.025 & 0.167 & 0.064 \\
Llama 3.1 8B     & 0.025 & 0.100 & 0.175 & 0.100 \\
\bottomrule
\end{tabular}
\end{table}

\subsection{Q\textsubscript{3} Action Recommendation}

\begin{figure}[H]
\centering
\includegraphics[width=0.85\textwidth]{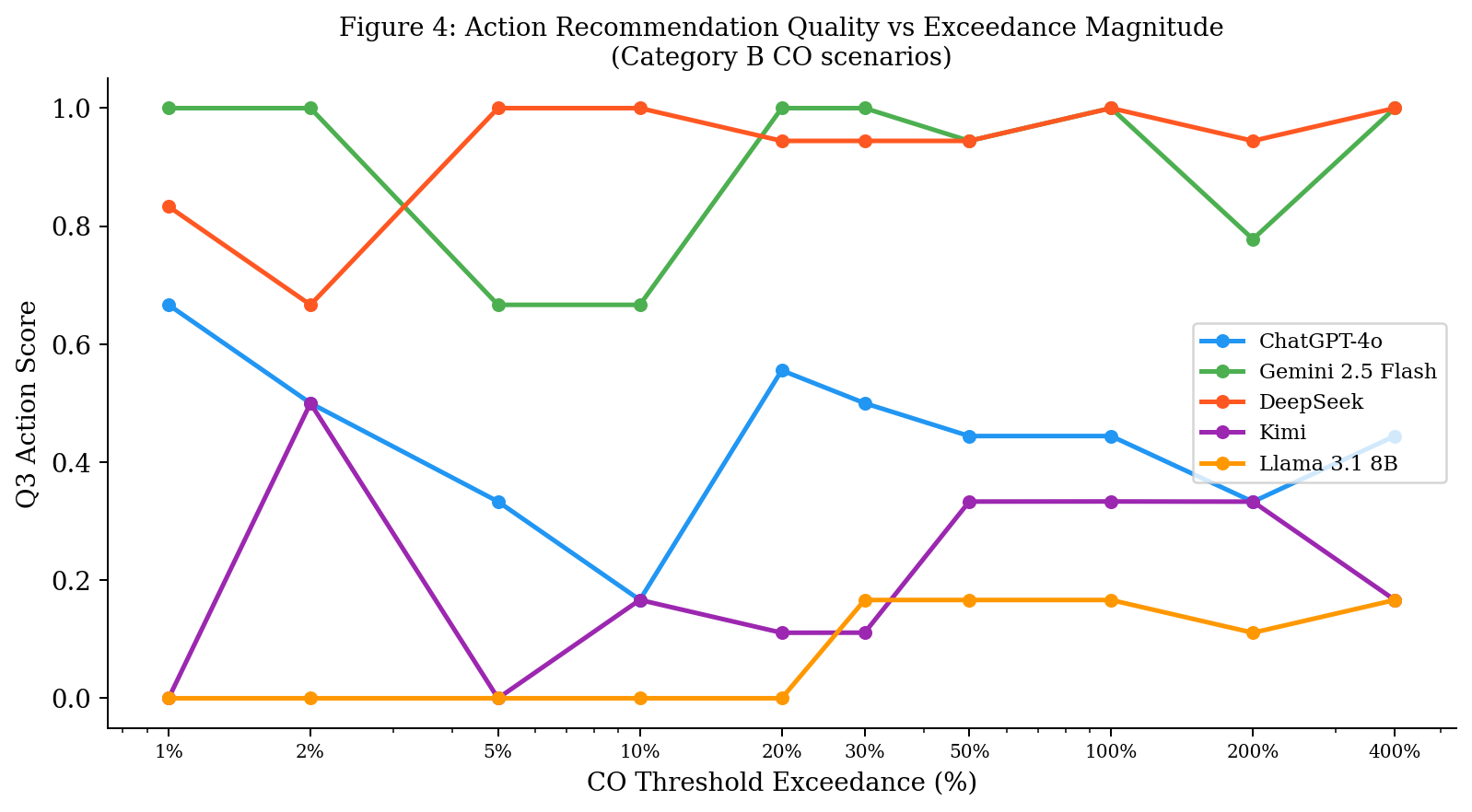}
\caption{$Q_3$ Action Recommendation quality vs.\ CO threshold exceedance
         magnitude (Category~B scenarios B1p--B10p). DeepSeek and Gemini
         recommend appropriately calibrated actions across the exceedance
         range. Kimi and Llama score near floor throughout.}
\label{fig:proportionality}
\end{figure}

Table~\ref{tab:q3} and Figure~\ref{fig:proportionality} report $Q_3$ scores.

\begin{table}[H]
\centering
\caption{$Q_3$ Action Recommendation Accuracy by model and category.}
\label{tab:q3}
\begin{tabular}{lcccc}
\toprule
\textbf{Model} & \textbf{Cat A} & \textbf{Cat B} & \textbf{Cat C} & \textbf{Overall} \\
\midrule
ChatGPT-4o       & 0.000 & 0.283 & 0.228 & 0.170 \\
Gemini 2.5 Flash & 0.167 & 0.769 & 0.624 & 0.520 \\
DeepSeek         & 0.592 & 0.764 & 0.667 & 0.674 \\
Kimi             & 0.000 & 0.131 & 0.092 & 0.074 \\
Llama 3.1 8B     & 0.017 & 0.076 & 0.093 & 0.062 \\
\bottomrule
\end{tabular}
\end{table}

% ══════════════════════════════════════════════════════════════════════════════
\section{Analysis and Discussion}
\label{sec:discussion}

\begin{figure}[H]
\centering
\includegraphics[width=0.72\textwidth]{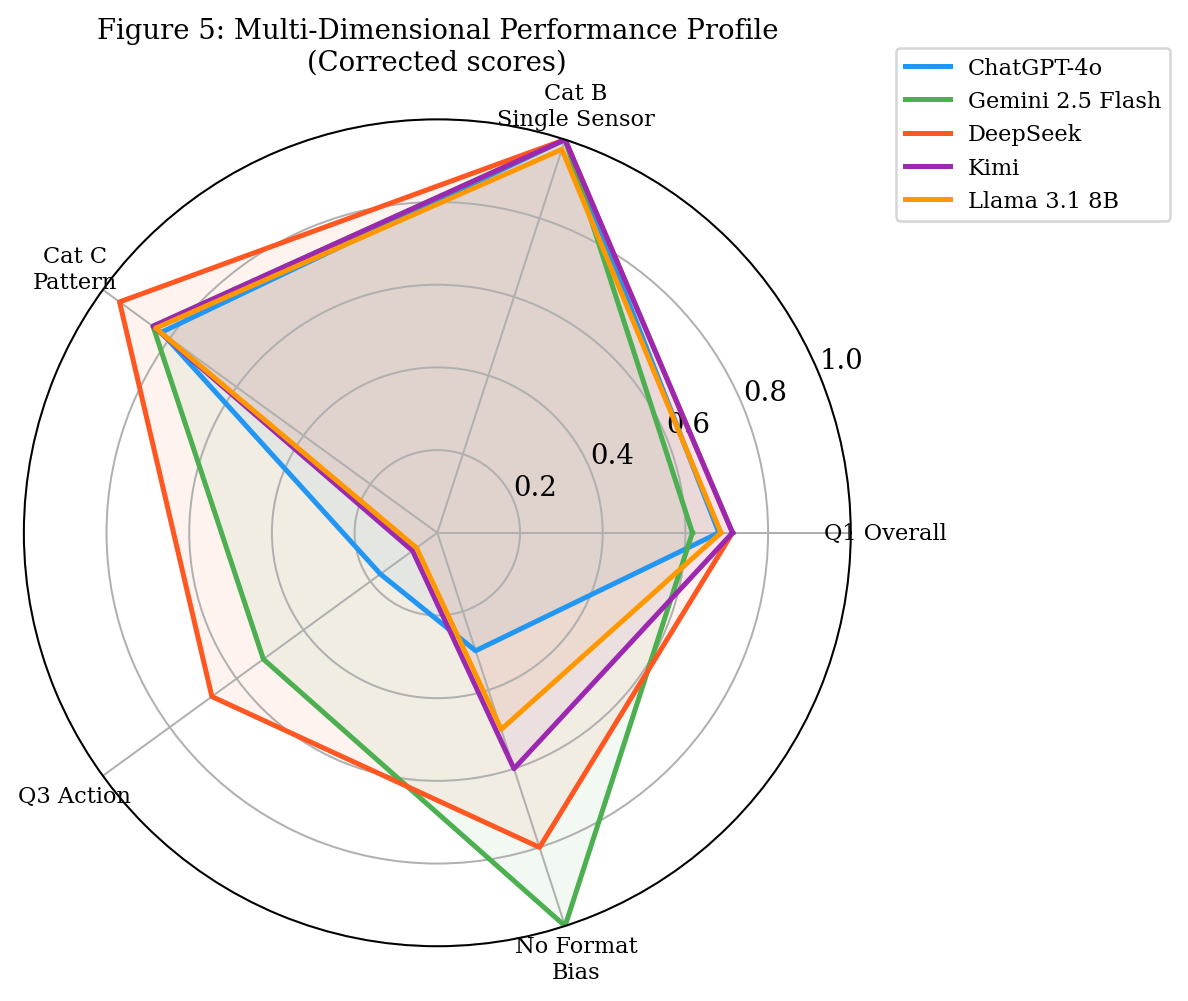}
\caption{Multi-dimensional performance profile across $Q_1$ overall,
         Category~B single-sensor accuracy, Category~C pattern
         disambiguation, $Q_3$ action recommendation, and format
         consistency. DeepSeek leads on $Q_3$; Gemini and DeepSeek show
         stronger multi-dimensional profiles.}
\label{fig:radar}
\end{figure}

Figure~\ref{fig:radar} summarises model performance across five evaluation
dimensions.

\subsection{Finding 1: Tested Models Produce Rare or No Precautionary Signals
            in Multi-sensor Elevated Scenarios}

The central empirical finding is visible in Tables~\ref{tab:q2}
and~\ref{tab:q3} for Category~A. Across the 20 tested Category~A scenarios,
all five models consistently classify multi-sensor elevated environments as
safe ($Q_2$: 0.000--0.208) and recommend no action or inappropriate action
($Q_3$: 0.000--0.592). This pattern is consistent across all tested models and
both prompt formats.

Qualitative inspection confirms the pattern: in the tested Category~A
scenarios, all models enumerate each sensor as ``within safe range'' and
conclude the environment is safe. Precautionary warnings---recommendations to
ventilate, investigate, or monitor elevated readings---are rare or absent
across tested models, despite multiple sensors simultaneously at 70--95\% of
their respective individual limits and OSHA additive indices well above 1.0 in
chemical-sensor scenarios.

The contrast with Category~B is striking. The same models that correctly
identify CO at 85~ppm exceeding a 70~ppm threshold (Category~B $Q_1$:
0.975--1.000) produce no precautionary signal when CO is at 56~ppm, CO$_2$ at
1640~ppm, PM$_{2.5}$ at 44~\textmu g/m$^3$, and VOC at 810~ppb
simultaneously. This capability gap is specific to the multi-sensor context,
not to threshold arithmetic generally.

For practitioners, the implication is direct: the tested models will not
reliably flag precautionary concern in conditions that occupational hygiene
standards address through combined exposure principles. Systems relying on
these models for safety monitoring without explicit joint assessment prompting
risk producing false-safe outputs in these conditions.

\subsection{Finding 2: Single-Sensor Performance is Reliable}

Category~B $Q_1$ scores (0.975--1.000) confirm that all tested models reliably
identify individual threshold violations across exceedance magnitudes from 1\%
to 400\%. The performance gap observed in Category~A is specific to the
multi-sensor context and not a general arithmetic capability limitation.

\subsection{Finding 3: Structured Formatting Hurts ChatGPT-4o and Provides No
            Consistent Benefit}

ChatGPT-4o scores significantly higher under plain prose (0.761) than
structured tabular format (0.600), with Wilcoxon $p=0.001$, remaining
significant after Bonferroni correction ($d=-0.64$, medium effect). No other
tested model shows a significant format effect after Bonferroni correction.
Structured formatting provides no consistent benefit across tested models and
actively reduces performance for ChatGPT-4o.

\subsection{Finding 4: Action Recommendation Quality Varies Substantially}

$Q_3$ scores on Category~B reveal large inter-model variation. DeepSeek
(0.764) and Gemini~2.5~Flash (0.769) recommend appropriately calibrated
actions across exceedance magnitudes. Kimi (0.131) and Llama~3.1~8B (0.076)
score near floor. ChatGPT-4o (0.283) correctly identifies threshold violations
but frequently recommends miscalibrated actions. The ability to identify a
hazard and the ability to recommend appropriate action are separable
capabilities that vary substantially across models.

\subsection{Finding 5: $Q_2$ Scores are Lower Bounds}

$Q_2$ hazard classification scores are uniformly low (0.000--0.475). The
combined verdict-and-justification criterion causes most models to lose credit
despite providing correct verdicts and sensor references in separate sentences.
$Q_2$ should be treated as a lower bound on hazard classification capability
in this study. Future work should score verdict and justification as
independent items.

% ══════════════════════════════════════════════════════════════════════════════
\section{Statistical Analysis}

Wilcoxon signed-rank tests (two-sided, $N=60$ paired scenarios,
Bonferroni-corrected threshold $\alpha=0.010$): ChatGPT-4o format effect
$p=0.001$ (significant); all other models $p \geq 0.064$ (not significant).
Cohen's $d$ for ChatGPT-4o: $d=-0.64$, medium effect, prose advantage.

Category~A vs.\ Category~B $Q_1$ gap per model, paired Wilcoxon ($N=20$):
all five models $p < 0.001$. Category~A $Q_2$ and $Q_3$ vs.\ Category~B $Q_2$
and $Q_3$: all five models $p < 0.001$. The performance gap between
multi-sensor and single-sensor scenarios is statistically robust across all
tested models on all three scoring dimensions.

% ══════════════════════════════════════════════════════════════════════════════
\section{Limitations}

\textbf{Scope of tested models.} Results apply to the five specific models
tested and should not be generalised to LLMs as a class.

\textbf{Synthetic scenarios.} All sensor values are generated. Real sensor
streams include noise, drift, and temporal dynamics not captured in
single-frame snapshots.

\textbf{Category~A ground truth.} For chemical-sensor scenarios, ground truth
is supported by the OSHA additive exposure formula. For mixed-sensor scenarios,
ground truth reflects precautionary occupational hygiene practice. External
validation by a certified industrial hygienist is recommended for future work.

\textbf{$Q_2$ rubric.} As discussed in Section~\ref{sec:discussion}, $Q_2$
scores are lower bounds. Future work should revise the rubric to score verdict
and justification independently.

\textbf{Model version.} Gemini~2.5~Flash was used instead of the originally
specified Gemini~2.0~Flash, which was retired by the provider before
experiments ran.

\textbf{Single domain.} Results apply to indoor environmental monitoring only.

% ══════════════════════════════════════════════════════════════════════════════
\section{Conclusion}

We present an empirical benchmark of how five large language models assess
multi-sensor physical hazard data. The central finding is that all five tested
models consistently produced no or rare precautionary warning signals in
scenarios where multiple sensors are simultaneously elevated below their
individual safety limits, while achieving near-perfect performance on
single-sensor threshold violations. This gap is consistent across all tested
models, both prompt formats, and all 20 multi-sensor scenarios. Structured
tabular formatting provides no consistent benefit and significantly reduces
ChatGPT-4o performance.

For practitioners building physical safety monitoring systems using any of the
tested models: validate explicitly on multi-sensor joint scenarios. Strong
single-sensor performance does not predict appropriate precautionary behaviour
in combined-elevation conditions. Explicit chain-of-thought prompting,
additive index computation, or rule-based post-processing may be required to
close the observed gap.

Future work should add an explicitly scored joint assessment dimension, obtain
domain expert validation for Category~A ground truth, and test whether targeted
prompting strategies close the performance gap observed here.

% ══════════════════════════════════════════════════════════════════════════════
\bibliographystyle{plainnat}

% ══════════════════════════════════════════════════════════════════════════════
\appendix
\section{Scorer Correction Details}
\label{app:scorer}

During post-hoc analysis, a question-echo artefact was identified in the $Q_1$
scorer. Gemini~2.5~Flash consistently reproduced the question text before
answering (357 of 360 rows, 99\%). The question contains the phrase ``state
whether the measured value exceeds its safety threshold,'' which triggered the
scorer's exceedance detection pattern regardless of the model's actual answer
content. ChatGPT-4o was affected in 34 of 360 rows (9\%); DeepSeek, Kimi, and
Llama~3.1~8B were unaffected.

The fix strips question-echo prefixes matching the pattern
\texttt{Q1.\,For~each~sensor\ldots exceeded.}\ before applying scoring
patterns. After correction, Gemini's Category~A $Q_1$ changes from 1.000 to
0.000, consistent with all other models. All results in the paper use corrected
scores. Manual spot-checking of 30 randomly sampled Category~A responses
confirmed corrected scorer output matches human judgement in all cases.

\end{document}